\documentclass[letterpaper]{article} % DO NOT CHANGE THIS
\usepackage{aaai2026}  % DO NOT CHANGE THIS
\usepackage{times}  % DO NOT CHANGE THIS
\usepackage{helvet}  % DO NOT CHANGE THIS
\usepackage{courier}  % DO NOT CHANGE THIS
\usepackage[hyphens]{url}  % DO NOT CHANGE THIS
\usepackage{graphicx} % DO NOT CHANGE THIS
\usepackage{natbib}  % DO NOT CHANGE THIS AND DO NOT ADD ANY OPTIONS TO IT
\usepackage{caption} % DO NOT CHANGE THIS AND DO NOT ADD ANY OPTIONS TO IT
\usepackage{algorithm}
\usepackage{algorithmic}

\usepackage{amssymb}
\usepackage{amsmath}
\usepackage{booktabs}

\usepackage{newfloat}
\usepackage{listings}
\DeclareCaptionStyle{ruled}{labelfont=normalfont,labelsep=colon,strut=off} % DO NOT CHANGE THIS
\floatstyle{ruled}
\newfloat{listing}{tb}{lst}{}
\floatname{listing}{Listing}
\title{Hybrid-Domain Adaptative Representation Learning for Gaze Estimation}
\author{
    Qida Tan\textsuperscript{\rm 1}, Hongyu Yang\textsuperscript{\rm 2}, Wenchao Du\textsuperscript{\rm 2}\thanks{Corresponding author}
}
\affiliations{
    \textsuperscript{\rm 1} National Key Laboratory of Fundamental Science on Synthetic Vision, Sichuan University, Chengdu, China\\
    \textsuperscript{\rm 2} College of Computer Science, Sichuan University, Chengdu, China\\
}

\usepackage{bibentry}
\begin{document}

\maketitle

\begin{abstract}
Appearance-based gaze estimation, aiming to predict accurate 3D gaze direction from a single facial image, has made promising progress in recent years. However, most methods suffer significant performance degradation in cross-domain evaluation due to interference from gaze-irrelevant factors, such as expressions, wearables, and image quality. To alleviate this problem, we present a novel Hybrid-domain Adaptative Representation Learning (shorted by HARL) framework that exploits multi-source hybrid datasets to learn robust gaze representation. More specifically, we propose to disentangle gaze-relevant representation from low-quality facial images by aligning features extracted from high-quality near-eye images in an unsupervised domain-adaptation manner, which hardly requires any computational or inference costs. Additionally, we analyze the effect of head-pose and design a simple yet efficient sparse graph fusion module to explore the geometric constraint between gaze direction and head-pose, leading to a dense and robust gaze representation. Extensive experiments on EyeDiap, MPIIFaceGaze, and Gaze360 datasets demonstrate that our approach achieves state-of-the-art accuracy of $\textbf{5.02}^{\circ}$ and $\textbf{3.36}^{\circ}$, and $\textbf{9.26}^{\circ}$ respectively, and present competitive performances through cross-dataset evaluation. The code is available at https://github.com/da60266/HARL.
\end{abstract}

%% Uncomment the following to link to your code, datasets, an extended version or similar.
%% You must keep this block between (not within) the abstract and the main body of the paper.
%\begin{links}
%    \link{Code}{https://aaai.org/example/code}
%    \link{Datasets}{https://aaai.org/example/datasets}
%    \link{Extended version}{https://aaai.org/example/extended-version}
%\end{links}

\section{Introduction}
\begin{figure}
	\centering
	\includegraphics[width=0.87\linewidth]{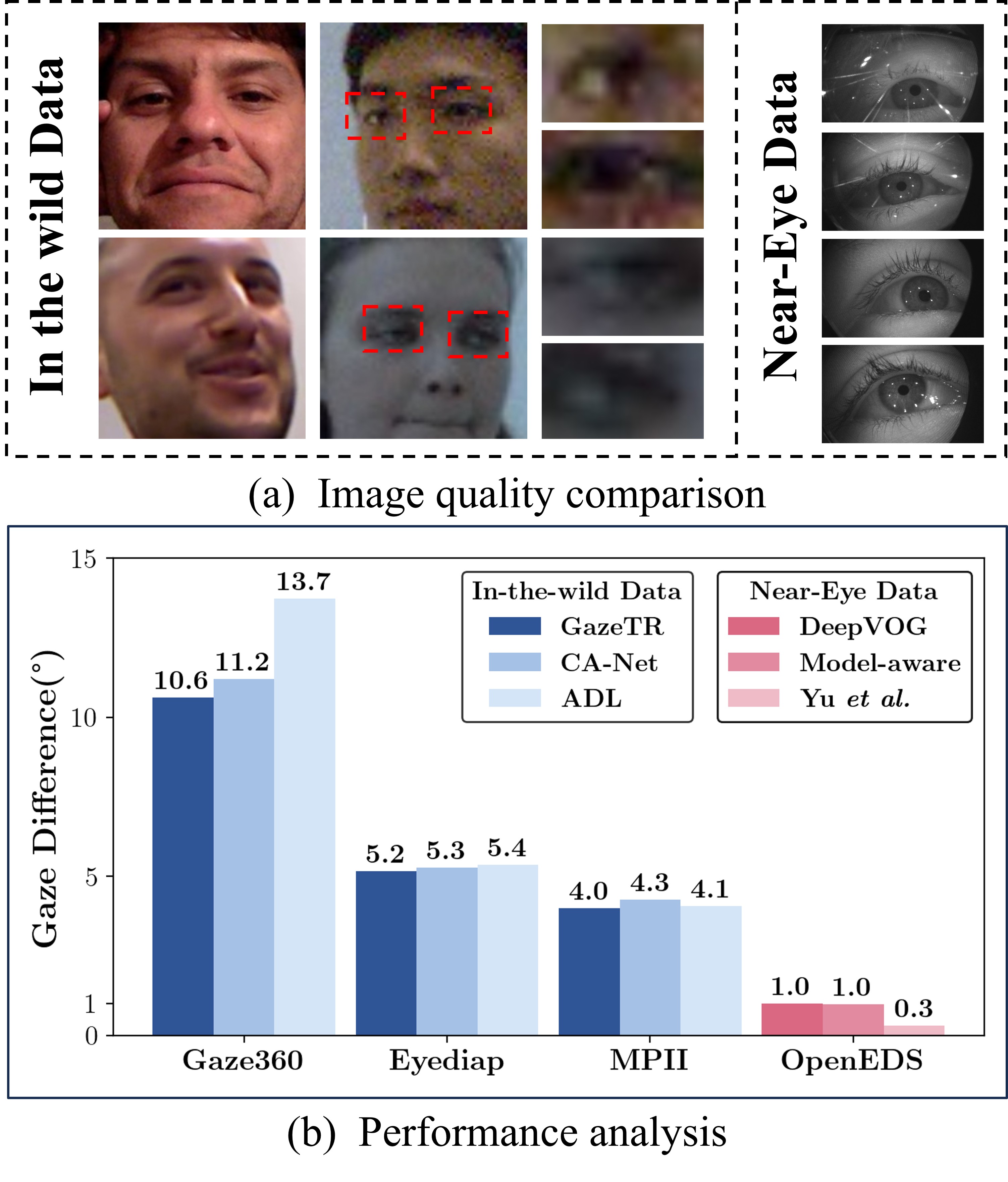}
	\caption{(a) Sample quality comparison between in-the-wild and near-eye datasets. (b) Performance comparisons between in-the-wild \cite{cheng2022gaze, cheng2020coarse, kellnhofer2019gaze360} and near-eye datasets\cite{yiu2019deepvog, popovic2023model,feng2022real} respectively.}
	\label{fig1}
\end{figure}

Gaze estimation, aiming to determine where someone is looking toward or visual attention is located, is a crucial clue for understanding human behaviors, and offers significant assistance for many practical applications, e.g., human-system interaction\cite{steil2018fixation}, mental fatigue detection\cite{christian2023pilot} and AR/VR systems\cite{bao2023exploring}. Existing methods can be approximately divided into two categories: geometry-based and appearance-based. The former focuses on the traditional image processing technologies and geometric gaze model to calculate the gaze direction, which only adapt to near-eye conditions and requires expensive hardware for high-resolution eyeball capturing. With the development of deep learning technologies, appearance-based methods achieve remarkable progress on gaze direction regression from the single facial image in recent years. However, a critical challenging lie in that the eye area only occupies a very small part of the whole face, which leads to the insufficient gaze cues due to lower resolution, e.g., geometric shapes and profiles of pupil and iris, as shown in Fig~\ref{fig1}(a). Furthermore, facial expressions, illumination and so on, also degrades the image quality, and leads to limited prediction accuracy and poor generalization on cross-domain evaluation.

Recently, all kinds of methods have been explored to alleviate the above problems, including feature distance minimization \cite{wang2022contrastive}, disentangling representation \cite{sun2021cross,yin2024clip}, and data generation \cite{ververas20253dgazenet}. However, a key point is often overlooked in them, which is that degraded eyeball appearances limit high-quality gaze representation, leading to poor generalization in cross-object gaze estimation. As illuminated in Fig~\ref{fig1}(b), high-resolution near-eye images result in the significantly higher prediction accuracy, even with a simple deep network. Therefore, exploiting the high-quality near-eye data to guide the model extracting gaze-relevant representation from low-quality facial image is valuable. However, directly applying existing domain-adaptation methods to aligning them in latent space is unrealistic due to the great domain gap, where the high- and low-quality monocular images are unpaired. Moreover, monocular gaze labels are always lost for low-quality facial images.

Inspired by the unsupervised domain-adaptation (UDA) regression, in this paper, we attempt to introduce high-quality monocular images to disentangle gaze-relevant representation from low-quality facial image with unsupervised learning. We present a novel Hybrid-domain Adaptative Representation Learning (HARL) framework that exploits the labeled high-quality near-eye data to extract monocular gaze representation from unlabeled low-quality facial images in a UDA manner. Specifically, HARL aligns the inverse Gram matrix of the hybrid-domain features to capture inner correlations, which is simple yet efficient and hardly requires any extra computation costs during training and inference. Furthermore, we also consider the effect of head-pose, and construct a sparse-graph fusion module to explore the latent geometry constraints between monocular gaze and head-pose representations, which leads to a dense and robust gaze representation. In short, the main contributions of the paper are summarized:

1) Propose an end-to-end gaze representation learning framework, i.e. HARL, which integrates the idea of UDA into the general appearance learning architecture to extract dense and robust gaze representation from low-quality facial images. \textit{As far as we know, it is the first UDA framework to disentangle gaze representation from hybrid-domain data,} and presents superior performances on in-domain and cross-domain evaluation.

2) Design a simple yet effective sparse-graph fusion module that explores the inherent geometric constraints between the monocular and pose features, and leads to the dense and robust facial gaze representation.

3) Extensive experiments demonstrate that the proposed HARL achieves state-of-the-art gaze accuracy of $\textbf{3.36}^{\circ}$, $\textbf{5.02}^{\circ}$ and $\textbf{9.26}^{\circ}$ on MPIIFaceGaze, EyeDiap and Gaze360 benchmarks, respectively, and also presents competing performances on cross-domain evaluations without any computational costs.

\section{Related Work}

\subsection{Appearance-based Gaze Estimation}
The early approaches focus on reconstructing the geometric structure of eyeball, which generally rely on the image processing technologies to locate the boundaries of pupil and iris. Therefore, the deep-learning-based eye segmentation is explored to support more accurate gaze prediction \cite{yiu2019deepvog, popovic2023model}. However, these methods achieved remarkable accuracy but also require personal calibration and dedicated devices such as depth sensors, infrared cameras and lights. Appearance-based approaches directly estimate gaze vector from the facial image captured by the web camera, which builds an end-to-end mapping between the image and the gaze label \cite{kellnhofer2019gaze360,wang2023gazecaps,o2022self}. Therefore, these methods have made great progress in recent years. However, a crucial challenge lies that the gaze representation is sensitive to facial appearances, which limits the prediction accuracy and generalization ability of the model. Leading works focus on exploring domain-adaptation (DA) \cite{bao2022generalizing,liu2021generalizing, Cai_2023_CVPR} and domain-generalization (DG) \cite{xu2023learning,bao2024feature,bao2024unsupervised,xu2024gaze} gaze estimation. The former aims to align gaze representations between the source and target domains, which generally requires to accessing the source and target domain data. Instead, the latter directly learn robust gaze representation from source domain data only. Although all of them achieved some improvements on cross-datasets evaluation, complex networks and carefully-designed training strategies leads to expensive computation costs, which is still unsuitable in practical applications. 
\begin{figure*}[t]
	\centering
	\includegraphics[width=\linewidth]{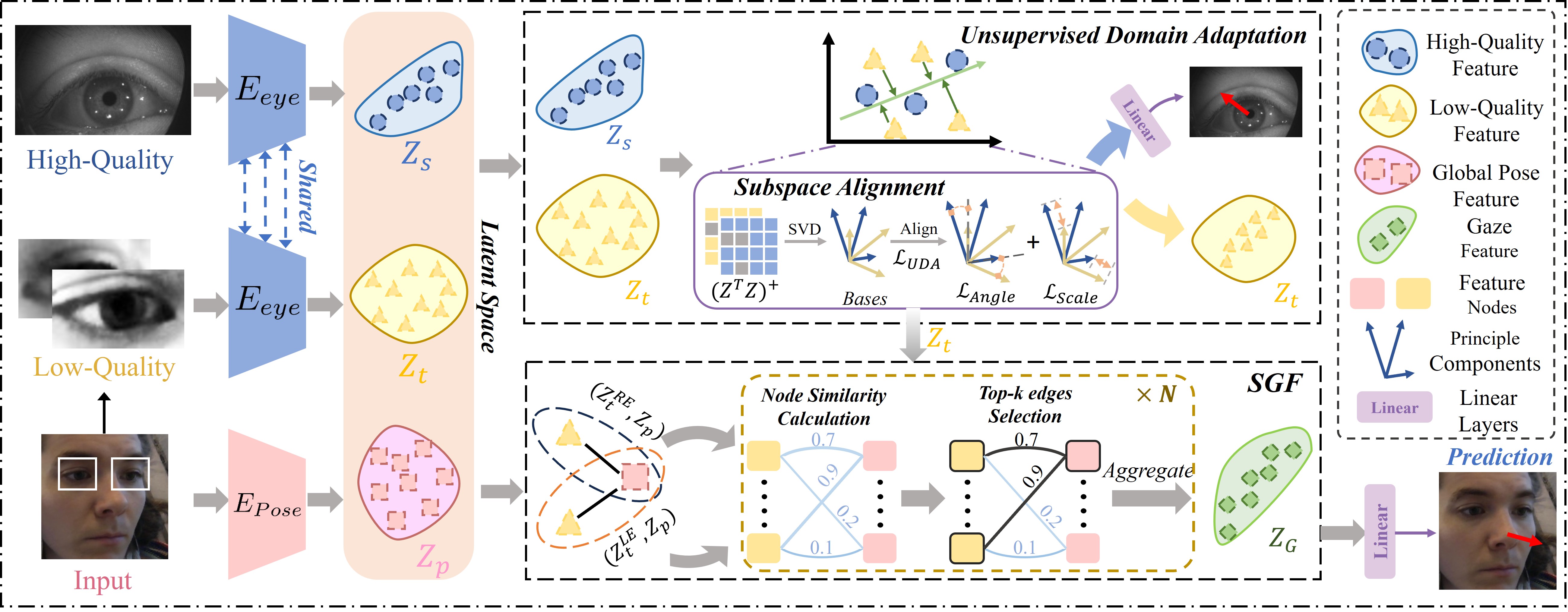}
	\caption{Overview of our framework. The proposed method explores a dual branch network to learning monocular and pose representation, a sparse-graph fusion module is used to generate dense gaze representation.}
	\label{fig2}
\end{figure*}
\subsection{Unsupervised Domain Adaptation}
Unsupervised domain adaptation (UDA) aims to adapt the model to target domain with unlabeled samples, which is widely applied to classical computer vision tasks \cite{rahman2020minimum,vu2019advent}. The general methods focus on feature alignment by adversarial learning or explicit losses such as maximum mean discrepancy (MMD) \cite{Long2017ConditionalAD,rahman2020minimum}, which have been applied to gaze estimation \cite{guo2020domain,wang2022contrastive}. However, these methods ignored a crucial problem that the robust gaze representation is difficult to acquire in the source domain due to the undesired degradation. Instead, our approach aims to disentangle the monocular gaze representation via UDA, which exploits the high-quality data as the source domain, low-quality facial data as the target domain, to capture feature correlations. It brings significant gains for in-dataset and cross-dataset evaluation.
\subsection{Graph Neural Networks}
Graph neural networks (GNNs) have received tremendous attention in causal reasoning from structured and non-structured data \cite{wu2020comprehensive}. The major component of the GNNs is the node feature aggregation technique, with which node can update its weight by interacting with other nodes \cite{kipf2017gcn}. Meanwhile, they adopt the same strategy in aggregating the information from different feature dimensions. However, inspired by recent advances on GNNs, there are potential benefits to treat the dimensions differently during the aggregation process \cite{jin2021gating}. Considering that extracting dense and robust gaze representations from low-quality images is challenging even with powerful supervision, thus, we investigate to enable heterogeneous contributions of feature dimensions with GNNs, which aims to explore fine-grained feature correlations for robust gaze representation. \textit{As far as we know, this is also the first work to exploit GNNs in gaze estimation.}

\section{Proposed Method}
Given a training sample $(I, g)$ drawn from specific domain $\mathcal{D}$, where $I$ denotes the input facial image, $g$ is the corresponding gaze label, which is generally expressed by the Euler angle (pitch and yaw) $(\phi, \psi)$. The appearance-based gaze estimation is defined as:
\begin{equation}
	Z = E(I;\Theta), \hat{\mathcal{G}} = R(Z;\beta),
	\label{eq1}
\end{equation}
where $E(\cdot)$ is an encoder, which aims to extract gaze-relevant representation $Z$ from input, $R(\cdot)$ represents a linear regressor that estimate the gaze vector $\hat{\mathcal{G}}$. Generally, $E$ and $R$ are coupled into an end-to-end appearance learning architecture. $\Theta$ and $\beta$ denote the learnable parameters.

In order to learn robust gaze representation $Z$, most appearance-based approaches, including supervised and unsupervised-based, focus on exploring more powerful encoder $E$ to minimize the distribution difference between source feature $Z_s$ and target feature $Z_t$, where the samples from source and target domains are all degraded facial images. In contrast, our method takes the high-resolution near-eye data as the source domain, the low-quality facial images are used as target domain. The key to it is aligning $Z_t$ to $Z_s$ to ensure effective monocular gaze representation learning, and then the head-pose factor is also considered with a novel sparse-graph fusion module, which leads to a dense and robust facial gaze representation. The proposed framework is shown in Fig~\ref{fig2}, which exploits a dual-branch encoding architecture, and contains two main modules, i.e. a Unsupervised Domain Adaptation Learning (UDAL) module, a Head-Pose injected Sparse-Graph Fusion (SGF) module.

\subsubsection{Unsupervised Domain Adaptation Learning}
Monocular gaze direction is a critical clue to infer facial gaze estimation. However, estimating it from low-quality facial images is challenging due to lack of monocular gaze labels and clear eye-appearance context. Therefore, UDAL exploits the idea of UDA to disentangle monocular gaze representation from low-quality facial images, which takes the high-quality monocular near-eye data as source domain, and leverages the pseudo-inverse low-rank property to align the scale and angle in a selected subspace generated by the pseudo-inverse Gram matrix of the two domains, leading to a disentangling representation learning framework.

Assuming that a linear regressor $R$ with parameter $\beta$ is utilized to estimate the monocular gaze directions $\mathcal{\hat{G}}$ in Eq~\ref{eq1}, i.e. $\mathcal{\hat{G}} = Z\beta$, which has an ordinary least squares (OSL) closed-form solution,
\begin{equation}
	\hat{\beta} = \left(Z^{T}Z\right)^{-1}Z^{T}\mathcal{\hat{G}}
	\label{eq2}
\end{equation}
where $\left(Z^{T}Z\right)^{-1}\in\mathbb{R}^{n\times n}$ is the inverse of the Gram Matrix. $Z^{T}\mathcal{\hat{G}}\in\mathbb{R}^{n\times 2}$ projects gaze feature to label space.

Suppose that we have a set of low- and high-quality near-eye image features with the same gaze direction. Our goal is to ensure that the low-quality image features would produce the same gaze direction through a shared $R$, thereby achieving consistent constraints between low- and high-quality image features, i.e. $Z_t$ and $Z_s$. This objective is formulated as:
\begin{equation}
	\begin{aligned}
		&\text{If } \mathcal{\hat{G}}_s = \mathcal{\hat{G}}_t \\
		&\text{then} \left(Z^{T}_{s}Z_{s}\right)^{-1}Z^{T}_{s}\mathcal{\hat{G}}_{s} = \left(Z^{T}_{t}Z_t\right)^{-1}Z^{T}_{t}\mathcal{\hat{G}}_{t},
	\end{aligned}
	\label{eq3}
\end{equation}
where $\mathcal{\hat{G}}_s$ and $\mathcal{\hat{G}}_t$ denote the gaze vectors from source and target images. When both them produce the same gaze prediction through a shared $R(\cdot;\beta)$, it implies that the encoder $E(\cdot)$ can extract consistent gaze-relevant features from different domains. By achieving it, \textit{low-quality gaze features $Z_t$ can be aligned with high-quality features $Z_s$, ensuring that the gaze representations remain highly relevant and contain rich gaze information.} Returning to the Eq~\ref{eq3}, it could be implemented by aligning $(Z^TZ)^{-1}Z^T$. Inspired by the recent advances on the UDA in regression \cite{Chen2021RepresentationSD,nejjar2023dare}, we can achieve the alignment of $(Z^TZ)^{-1}Z^T$ by incorporating angular and scale constraints on the inverse Gram matrix $(Z^TZ)^{-1}$ during training, which not only ensures the alignment of the inverse Gram matrix but also contributes to a well-aligned $Z$.

The Gram Matrix $Z^{T}Z$ of the feature can be decomposed using singular value decomposition (SVD) \cite{gloub1996matrix}:
\begin{equation}
	\left(Z^{T}Z\right)=\left(UDV^{T}\right)^{T}\left(UDV^{T}\right)=V \Lambda V^{T},
	\label{eq4}
\end{equation}
where the orthogonal matrix $V \in \mathbb{R}^{n\times n}$ is identical to the matrix in the SVD of $Z$ and $\Lambda\in \mathbb{R}^{n\times n}$ is the diagonal matrix containing the squared eigenvalues of $Z$. Given a feature matrix $Z\in \mathbb{R}^{b\times n}$ where $n$ is always greater than $b$, the corresponding gram matrix has rank $r\leq b$. As a result, the Gram matrix is not full-rank and therefore not invertible. In such cases, the Moore-Penrose-pseudo-inverse \cite{ben2006generalized} can be used to generalize the concept of the matrix inverse and provide a stable solution for further computations involving matrix inversion. Given the ordered eigenvalues of the $\left(Z^{T}Z\right) \in \mathbb{R}^{n\times n}$, $\lambda_1\  \geq...\geq \lambda_r \geq...\geq \lambda_{n}$, the pseudo-inverse of $\left(Z^{T}Z\right)$ can be expressed as:
\begin{equation}
	\begin{aligned}
		G^{+} &= \left(Z^{T}Z\right)^{+} = V \Lambda^{+} V^{T} \\
		&=
		V\left(
		\begin{array}{c|c}
			\begin{array}{ccc}
				\frac{1}{\lambda_1} & &  \\
				& \ddots &  \\
				& & \frac{1}{\lambda_r}
			\end{array}
			&
			\begin{array}{c}
				0 \\
				\vdots \\
				0
			\end{array} \\ \hline
			\begin{array}{ccc}
				0 & \cdots & 0
			\end{array}
			&
			0
		\end{array}
		\right)
		V^{T}.
	\end{aligned}
	\label{eq5}
\end{equation}

The $r$-principal components derived from the the $\Lambda$ can be regarded as the subspace bases. Then we can measure the principle angles of source and target spaces using the following equation:
\begin{equation}
	\cos\left(\theta_{i}^{s\leftrightarrow t}\right)= \frac{G^{+}_{i,s} \cdot G^{+}_{i,t}}{\vert\vert{G^{+}_{i,s}}\vert\vert\cdot \vert\vert{G^{+}_{i,t}}\vert\vert}.
	\label{eq6}
\end{equation}

The cosine similarity between the spans of the source and target subspaces is given by $M=\left[\cos\left(\theta^{s\leftrightarrow t}_{1}\right),...,\cos\left(\theta^{s\leftrightarrow t}_{n}\right)\right]$. We aim to make the angles between the two subspaces as close as possible, leading to the following angle alignment constraint:
\begin{equation}
	\mathcal{L}_{Angle}\left(Z_{s},Z_{t}\right)=\vert\vert\mathbb{I}-M\vert\vert_{1},
	\label{eq7}
\end{equation}
where $\mathbb{I}$ is the vector of $1$ with shape of $n$.
The scale alignment is regularized by minimizing the difference between the $r$-principal eigenvalues,
\begin{equation}
	\mathcal{L}_{Scale}\left(Z_{s},Z_{t}\right)=\vert\vert\ \lambda_{s,i=1,..,r}-\lambda_{t,i=1,..,r}\vert\vert_{2}.
	\label{eq8}
\end{equation}

Finally, the Unsupervised Domain-Adaptation (UDA) loss is the combination of these two items,
\begin{equation}
	\mathcal{L}_{\textit{UDA}}=\mathcal{L}_{Angle}\left(Z_{s},Z_{t}\right) + \mathcal{L}_{Scale}\left(Z_{s},Z_{t}\right).
	\label{eq9}
\end{equation}

As shown in Fig~\ref{fig2}, in practice, we first crop the eye regions from the facial image and combine them with high-quality monocular near-eye images as input to the shared encoder $E_{\text{eye}}$, it hardly requires any extra computational costs.

\subsubsection{Sparse-Graph Fusion}
Previous works \cite{yue2024gaze, liang2025confounded} have explored the effect of head-pose for gaze estimation, it heavily affects the prediction accuracy in the wild. Existing methods directly model a global appearance-based network to extract pose prior from the facial image, which always leads to poor generalization due to the irrelevant interference (e.g., facial texture and expression). To this end, we also consider the effect of the head-pose but exploit a more simple and effective manner. We take a pretrained dense facial landmark model to extract robust pose representations, which provides rich pose and gaze clues by landmarks locations. This process is defined as
\begin{equation}
	Z_p = E_{Pose}(I;\theta),
	\label{eq10}
\end{equation}
where the $\theta$ is fixed model parameter and $I$ denotes the input image. In practice, we use the output of the penultimate layer of the model as dense pose representation $Z_{p}$.

Considering the robust facial gaze representation is always low-ranked and dense. After extracting binocular gaze features $Z_t = \{Z^{LE}_t, Z^{RE}_t\}$ and pose feature $Z_p$, where $Z^{LE}_t$ and $Z^{RE}_t$ denote the feature embeddings from the left- and right-eye images separately, we further explore the SGF module to capture the inner geometric constraints among them, while aggregating highly-relevant gaze information.

Supposing binocular features $\{Z^{LE}_t, Z^{RE}_t\}$ and facial pose feature $Z_p$ are viewed as the single node, separately, it is too simple to capture effective gaze features with three nodes and two edges in the resulting graph. Moreover, although the representations are low-dimensional and dense, they remain affected by interference-induced redundancy. Inspired by the efficient feature gating mechanism \cite{jin2021gating} that dynamically adjusts the contribution of each feature dimension during aggregation to enhance the impact of the important features, we further construct a subgraph between the monocular and pose features, where each feature dimension in them is treated as an individual node, and edges are defined based on feature similarity among each dimension. A two-layer MLP network is employed to compute the node similarity between monocular and facial pose features in a self-adaptive manner:
\begin{equation}
	\mathcal{S}_{i,j}=Linear(\sigma(Linear(n_{i} - n_{j}^{p})))
	\label{eq11}
\end{equation}
where $\sigma$ denotes a ReLU activation. In addition, $n$ and $n^p$ denote the nodes from binocular gaze and facial pose representations, respectively.

We select the top-$k$ pose feature nodes that related to each monocular feature node in $Z_t$. The adjacent matrix $\mathcal{AD}$ of all nodes in the single subgraph is constructed as
\begin{equation}
	\mathcal{AD}_{i,j} =
	\begin{cases}
		1, & \text{if } j\in \text{top-}k\{\mathcal{S}_{i,k}\mid k = 1, \dots, q \}, \\
		0, &\text{otherwise}
	\end{cases}
	\label{eq12}
\end{equation}
where $q$ indicates the number of dimensions.

To capture inherent geometric relationship for each node, we iteratively update node features by aggregating neighboring nodes at each layer,

\begin{equation}
	\begin{aligned}
		&n_{i}^{l+1} = Linear(n_{i}^{l}) + \sum_{j \in \mathcal{N}_i}Linear(n_j^{(l,p)})\\
		&\quad \text{where} \quad \mathcal{N}_i = \{ j \mid \mathcal{AD}_{i,j} = 1 \},
	\end{aligned}
	\label{eq13}
\end{equation}
where $\mathcal{N}_i$ denotes a set of neighborhood nodes with center node $n_i$, and the $l$ indicates the current layer of GNNs.

After aggregating node features through several GNNs layers, SGF output a unified full-face gaze representation $Z_{G}$, the final gaze vector regression is implemented by stacking two linear layers.

\subsubsection{Joint Optimization}
The proposed HARL is an end-to-end learning framework, the total loss function is defined as:
\begin{align}
	\mathcal{L}_{total} =
	\mathcal{L}_{\textit{MSE}}(\mathcal{G}_{Face}, \hat{\mathcal{G}}_{Face}) &+ \mathcal{L}_{\textit{MSE}}(\mathcal{G}_{eye}, \hat{\mathcal{G}}_{eye}) \notag \\
	& + \lambda\mathcal{L}_{\textit{UDA}}(Z_{s}, Z_{t})
	\label{eq15}
\end{align}
where the mean square error (MSE) is used to supervise gaze vector regression. $\mathcal{\hat{G}}_{Face}$ and $\hat{\mathcal{G}}_{eye}$ denote predicted gaze vector for facial and near-eye images separately. $Z_s$ and $Z_t$ are the extracted monocular gaze representations from high- and low-quality monocular images respectively. $\lambda$ is a hyperparameter to balance the effect of UDA loss.

\section{Experiments}
\label{sec4}
\subsection{Experimental Setup}
\subsubsection{Datasets.} We select three benchmarks to evaluate the performance of HARL, i.e. Eyediap \cite{funesmora2014eyediap}, MPIIFace \cite{zhang2019mpiifacegaze} and Gaze360 \cite{kellnhofer2019gaze360}. Eyediap has $16k$ images captured in a controlled laboratory environment with screen and floating targets. We divide it into four clusters and apply four-fold cross-validation for in-dataset tests. MPIIFace Consists of $45k$ images captured by webcams during daily laptop usage. We perform leave-one-subject-out cross-validation for in-dataset tests. Gaze360 comprises $101k$ images collected using a 360° camera in outdoor street settings. Considering our HARL depends on monocular gaze representations from left and right-eye images, we filtered out samples with missing eye regions due to large pose during the training. Furthermore, we use the OpenEDS2020 \cite{palmero2021openeds2020} as the high-quality source data, which contains $180k$ high-resolution eye images captured using head-mounted VR/AR devices. We only select a subset for training.

\subsubsection{Implementation Details.} We use the ResNet18 \cite{he2016deep} as the basic encoder to implement the UMGRL, and an extra linear layer is exploited to regress monocular gaze directions for high-quality near-eye data. For SGF module, we select the PiPNet \cite{Jin2020PixelinPixelNT} as the pretrained pose encoder, which also used the ResNet18 as the basic encoder.  Then we implement our sparse-graph network with 4 GNN layers. We use the SGD optimizer with learning rate $lr=1e^{-4}$ during training. High-resolution monocular images from OpenEDS2020 are center-cropped and resized to $224 \times224$. Low-quality monocular images are also cropped from the facial images, and then are resized and grayscaled. The hyperparameter $\lambda$ in Eq~\ref{eq15} is set 0.5. We train HARL with 50, 30, and 30 epochs on EyeDiap, MPIIFace, and Gaze360 respectively for in-domain evaluation. For cross-domain evaluation, we use Gaze360 as the source domain, and train the model with 20 epochs only.

For in-domain evaluation, we select some representative methods for comparison, including ADL \cite{kellnhofer2019gaze360}, SAtten-net \cite{o2022self}, GazeCaps \cite{wang2023gazecaps}, 3DGazeNet \cite{ververas20253dgazenet}, IEH \cite{yue2024gaze}, L2CS-NET \cite{abdelrahman2023l2cs}, GazeTR (including GazeTR-Pure and GazeTR-Hybrid) \cite{cheng2022gaze}, CA-Net \cite{cheng2020coarse} and GFNet \cite{hu2023gfnet}. For cross-domain evaluation, we also select representative Domain-Adaptation methods and Domain-Generalization (DG) methods for comprehensive comparison, including PnP-GA \cite{liu2021generalizing}, DAGEN \cite{guo2020domain}, ADDA \cite{tzeng2017adversarial}, CRGA \cite{wang2022contrastive}, GVBGD \cite{cui2020gradually}, RUDA \cite{bao2022generalizing}, Full-Face \cite{zhang2017s}, RT-Gene \cite{fischer2018rt}, CA-Net \cite{cheng2020coarse}, GazeTR \cite{cheng2022gaze}, GazeCaps \cite{wang2023gazecaps}, ADL \cite{kellnhofer2019gaze360}, PureGaze \cite{cheng2022puregaze}, CLIP-Gaze \cite{yin2024clip}, GazeCon \cite{xu2023learning}, AGG \cite{bao2024feature}, and GLA\cite{zeng2025gaze}. Moreover, The Mean Angular Error (MAE) is used as the common evaluation metric.
\begin{table}[t]
	\centering
	\setlength{\tabcolsep}{4.5pt}
	\begin{tabular}{l|ccc}
		\hline
		\textbf{Method} & \textbf{MPIIFace} & \textbf{EyeDiap} & \textbf{Gaze360} \\
		\hline
		ADL  & $4.06^{\circ}$ & $5.36^{\circ}$ & $13.73^{\circ}$ \\
		SAtten-net  & $4.04^{\circ}$ & $5.25^{\circ}$ & $10.70^{\circ}$ \\
		GazeCaps  & $4.06^{\circ}$ & \underline{$5.10^{\circ}$} & $10.40^{\circ}$ \\
		3DGazeNet  & $4.0^{\circ}$ & - & \underline{$9.60^{\circ}$} \\
		IEH  & \underline{$3.49^{\circ}$} & - & $9.89^{\circ}$ \\
		L2CS-NET  & $3.92^{\circ}$ & - & $10.40^{\circ}$ \\
		GazeTR-Pure  & $4.74^{\circ}$ & $5.72^{\circ}$ & $13.58^{\circ}$ \\
		GazeTR-Hybrid  & $4.00^{\circ}$ & $5.17^{\circ}$ & $10.62^{\circ}$ \\
		CA-Net  & $4.27^{\circ}$ & $5.27^{\circ}$ & $11.20^{\circ}$ \\
		GFNet  & $3.96^{\circ}$ & $5.40^{\circ}$ & -\\
		\hline
		\textbf{HARL} & $\textbf{3.36}^{\circ}$ & $\textbf{5.02}^{\circ}$ & $\textbf{9.26}^{\circ}$\\
		\hline
	\end{tabular}
	\caption{Mean Angular Error (MAE) results of different methods. The best and second-best results are \textbf{bolded} and \underline{underlined}, respectively.}
	\label{tab1}
\end{table}
\begin{table}[t]
	\centering
	\setlength{\tabcolsep}{2pt}
	\begin{tabular}{l|ccccc|c}
		\toprule
		Methods & Type& Source & Target & $\text{G}\rightarrow\text{E}$ & $\text{G}\rightarrow\text{M}$& Avg. \\
		\midrule
		PnP-GA  & DA & Yes & 100 & 7.92$^{\circ}$ & \underline{6.18$^{\circ}$}& 7.05$^{\circ}$ \\
		DAGEN  & DA & Yes & 500 & 12.90$^{\circ}$ & 8.74$^{\circ}$ &10.8$^{\circ}$\\
		ADDA  & DA & Yes & 500 & 12.90$^{\circ}$ & 6.61$^{\circ}$ & 9.76\\
		CRGA  & DA & Yes & 100 & \underline{6.68$^{\circ}$} & \textbf{6.09$^{\circ}$}& \underline{6.39$^{\circ}$}\\
		GVBGD  & DA & Yes & 1000 & 12.44$^{\circ}$ & 7.64$^{\circ}$ & 10.0$^{\circ}$\\
		RUDA  & DA & Yes & 100 & \textbf{5.86$^{\circ}$} & 6.20$^{\circ}$ & \textbf{6.03$^{\circ}$}\\
		\hline
		Full-Face  & DG & No & 0 & 14.42$^{\circ}$ & 11.13$^{\circ}$ &12.8$^{\circ}$\\
		RT-Gene  &  DG & No & 0 & 38.60$^{\circ}$ & 21.81$^{\circ}$ & 30.2$^{\circ}$\\
		CA-Net  &  DG & No  & 0 & 31.41$^{\circ}$ & 27.13$^{\circ}$ & 29.3$^{\circ}$\\
		GazeTR &  DG & No  & 0 & 8.88$^{\circ}$ & 7.96$^{\circ}$ & 8.42$^{\circ}$\\
		GazeCaps  & DG & No  & 0 & 9.20$^{\circ}$ & 9.20$^{\circ}$ & 9.20$^{\circ}$\\
		ADL  & DG & No  & 0 & 11.86$^{\circ}$ & 11.36$^{\circ}$ & 11.6$^{\circ}$\\
		PureGaze  & DG & No & 0 & 9.32$^{\circ}$ & 9.28$^{\circ}$ & 9.30$^{\circ}$\\
		CLIP-Gaze$^-$  & DG & No & 0 & 7.73$^{\circ}$ & 7.55$^{\circ}$ & 7.74$^{\circ}$\\
		CLIP-Gaze  & DG & No & 0 & \textbf{7.06$^{\circ}$} & \textbf{6.89$^{\circ}$} & \textbf{6.98$^{\circ}$}\\
		GazeCon  & DG & No & 0 & 8.52$^{\circ}$ & 7.82$^{\circ}$ & 8.17$^{\circ}$\\
		AGG  & DG & No & 0 & 7.93$^{\circ}$ & 7.87$^{\circ}$ & 7.90$^{\circ}$\\
		GLA  & DG & No & 0 & \underline{7.55$^{\circ}$} & 7.62$^{\circ}$ & 7.59$^{\circ}$\\
		\hline
		\textbf{HARL} & DG & No & 0 & $7.63^{\circ}$ & \underline{$7.49^{\circ}$} & \underline{7.56$^{\circ}$}\\
		\bottomrule
	\end{tabular}
	\caption{MAE results for cross-domain evaluations. 'Type' denotes whether the method belongs to domain-adaptation (DA) or domain-generalization (DG). 'Source' indicates whether the source domain sample is required during testing. 'Target' denotes whether the target-domain sample is required and the number of samples from the target domain. The 'G→E' and 'G→M' represents using Gaze360 as source domain, EyeDiap and MPIIFace are viewed as target domains, respectively.}
	\label{tab2}
\end{table}
\begin{table}[t]
	\centering
	\setlength{\tabcolsep}{3pt}
	\begin{tabular}{l|ccc|r}
		\toprule
		Variant & EyeDiap & G $\rightarrow$ M & G $\rightarrow$ E & \#Params  \\
		\midrule
		Baseline & $5.54^{\circ}$ & $11.97^{\circ}$ & $12.56^{\circ}$ & -- \\
		\midrule
		w/ $\mathcal{L}_{\textit{UDA}}$ & $5.42^{\circ}$ & $9.81^{\circ}$ & $9.49^{\circ}$ & + 0.M\\
		w/ $E_{\textit{Pose}}$ & $5.35^{\circ}$ & $9.69^{\circ}$ & $10.42^{\circ}$ & + 0.M \\
		w/ \textit{SGF} & $5.47^{\circ}$ & $9.48^{\circ}$ & $11.08^{\circ}$ & + 2.1M\\
		w/ $\mathcal{L}_{\textit{UDA}}$ + $E_{\textit{Pose}}$ & $5.20^{\circ}$  & $8.08^{\circ}$  & $8.48^{\circ}$ & + 0.M\\
		w/ $E_{\textit{Pose}}$ + \textit{SGF} &  $5.23^{\circ}$  & $8.35^{\circ}$  & $8.65^{\circ}$ & + 2.1M\\
		w/ $\mathcal{L}_{\textit{UDA}}$ + \textit{SGF} & $5.17^{\circ}$  & $8.27^{\circ}$ & $9.45^{\circ}$ & + 2.1M\\
		Full Model & $\textbf{5.02}^{\circ}$  & $\textbf{7.49}^{\circ}$ &   $\textbf{7.63}^{\circ}$ & + 2.1M\\
		\bottomrule
	\end{tabular}
	\caption{MAE results from each variant. The 'G→M' and 'G→E' denote using Gaze360 as source domain for training, MPIIFace and EyeDiap are viewed as target domains for evaluation, respectively.
	}
	\label{tab3}
\end{table}
\begin{table}[t]
	\centering
	\setlength{\tabcolsep}{7pt}
	\begin{tabular}{c|cccc}
		\toprule
		Strategy & w/o SubGraph & top-3 & top-2 & top-1 \\
		\midrule
		MAE      & $5.55^{\circ}$ & $5.22^{\circ}$  & $5.13^{\circ}$  & $\textbf{5.02}^{\circ}$\\
		\bottomrule
	\end{tabular}
	\caption{MAE results on EyeDiap with different graph construction strategies.}
	\label{tab4}
\end{table}
\begin{figure}[t]
	\centering
	\includegraphics[width=\linewidth]{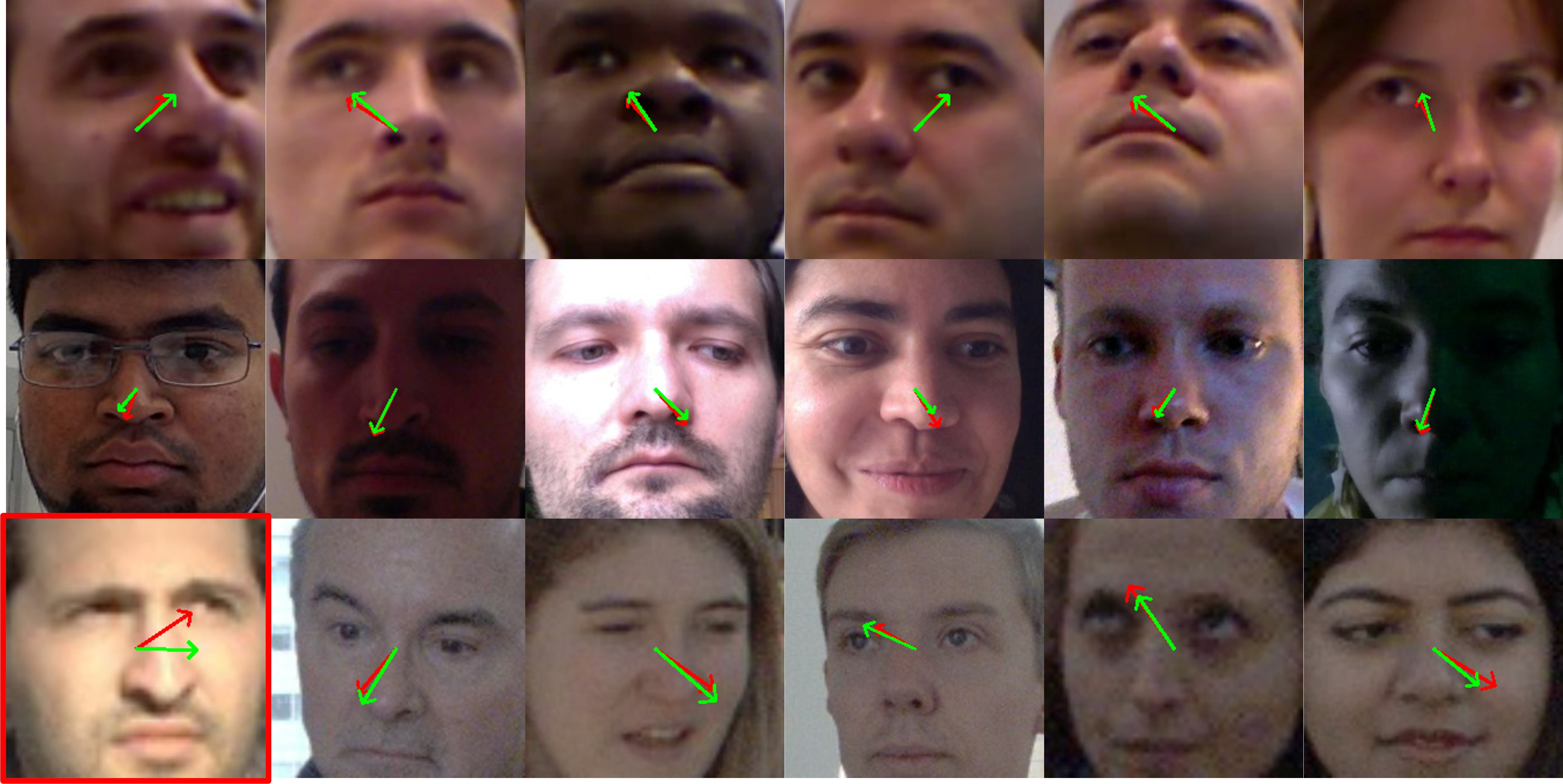}
	\caption{Visualized results from different datasets. The red and green arrows denote the predicted and ground truth gaze vector, separately.}
	\label{fig3}
\end{figure}
\subsection{Comparison with SOTA Methods}
\subsubsection{In-Domain Evaluation} 
We first perform in-domain evaluation on three benchmarks, and the quantitative results are listed in Tab~\ref{tab1}. Most methods have achieved some improvements on MPIIFace and EyeDiap, e.g., IEH \cite{yue2024gaze} also explored the transformer blocks to fuse local monocular and global facial gaze representation, which decreased the MAE to $3.49^{\circ}$ on MPIIFace. GazeCaps \cite{wang2023gazecaps} exploited the capsule network with a self-attention-routing mechanism, and achieved the competing result with $5.10^{\circ}$ on EyeDiap. Instead, our HARL achieves significantly better MAE results with $3.36^{\circ}$ and $5.02^{\circ}$ on MPIIFace and EyeDiap. Moreover, there is a clear tendency for all methods to experience significant performance degradation on Gaze360, mainly due to its more unconstrained data collection environment, which leads to greater degradation. Our method also achieves the best MAE with clear gains against 3DGazeNet \cite{ververas20253dgazenet}. It supports that our HARL could capture reliable gaze representations for low-quality facial images. Fig~\ref{fig3} gives the visualized results on typical real scenes, our method presents better robustness for lighting, expression, and pose. Note that the red rectangle denotes a special sample with an error gaze label, but HARL still predicts the correct gaze direction. 

\subsubsection{Cross-Domain Evaluation} 
Tab~\ref{tab2} lists the detailed results on the two cross-domain evaluation tasks. It is clear that the DA-based method achieved significantly better MAE results than the DG-based methods. However, they always require source and target samples to fine-tune the model, which is impossible in practical scenes. DG-based methods present competing performances, although with slightly higher MAEs, e.g., CLIP-Gaze \cite{yin2024clip} achieved the best MAE results with $7.06^{\circ}$ and $6.89^{\circ}$ on EyeDiap and MPIIFace benchmarks separately compared to other methods. However, CLIP-Gaze requires explicit text prompts corresponding to the facial image, e.g., expressions, illuminations, poses, and glasses. Removing personalized prompts, i.e. CLIP-Gaze$^-$, the performance drops a lot. In contrast, our HARL hardly requires any extra computational costs during the training and testing, which significantly improves the generalization ability with competing MAEs, making it more suitable for real-world applications.
\subsection{Ablation Study}
In this section, we mainly analyze the effect of key components in our HARL, including UDAL (i.e. $\mathcal{L}_{\textit{UDA}}$), $E_{\textit{Pose}}$ and SGF module, experiments are performed on EyeDiap and MPIIFace to evaluate their effects under in-domain and cross-domain settings.
\subsubsection{Component Analysis}
Our baseline exploits a dual branch network to encoder monocular gaze and global pose representations separately, which are simply implemented with two ResNet-18 networks, the outputs are concatenated along channel dimension and then fed into a two-layer MLP to predict final gaze direction. Note that the pretrained pose network also used the ResNet-18 as the backbone. Then we insert the $\mathcal{L}_{\textit{UDA}}$, $E_{\textit{Pose}}$, and SGF into the baseline, where the $\mathcal{L}_{\textit{UDA}}$ and $E_{\textit{Pose}}$ hardly increase any extra parameters and computational costs due to the shared encoder. 

Tab~\ref{tab3} lists the detailed results. The baseline present some ability on in-domain evaluation, but it has poor generalization on cross-domain tasks. Integrating $\mathcal{L}_{\textit{UDA}}$, $E_{\textit{Pose}}$ and SGF into baseline all bring some gains for in-domain and cross-domain evaluation, but w/ $\mathcal{L}_{\textit{UDA}}$ leads to significantly better generalization performances. Furthermore, the variants with different combinations among them all bring clear performance gains compared to the variants with single module only. Finally, our full model (i.e. HARL) achieves the most significant performance improvements across in-domain and cross-domain evaluations with few parameters.

\subsubsection{Sparse Graph Construction} For SGF module, we also explore the effect of sparse subgraph construction, the results are shown in Tab~\ref{tab4}. Firstly, removing subgraph in each layer (i.e., the monocular gaze features and pose feature are viewed as individual node) leads to a more simple graph with three nodes and two edges only, it results in the obvious performance drop, i.e. inner geometric constraints among them are not exploited well. Furthermore, we explore the sparse edge connection in subgraph, which relies on adjacent matrix $\mathcal{AD}$ in Eq~\ref{eq12}. We select different top-$k$ to construct subgraph, and observe an interesting phenomenon that using fewer similar nodes to construct edges brings the better performances, which leads to an extremely sparse subgraph in each layer of SGF. This also significantly enhances the inference efficiency of the model.

\begin{figure}
	\centering
	\includegraphics[width=\linewidth]{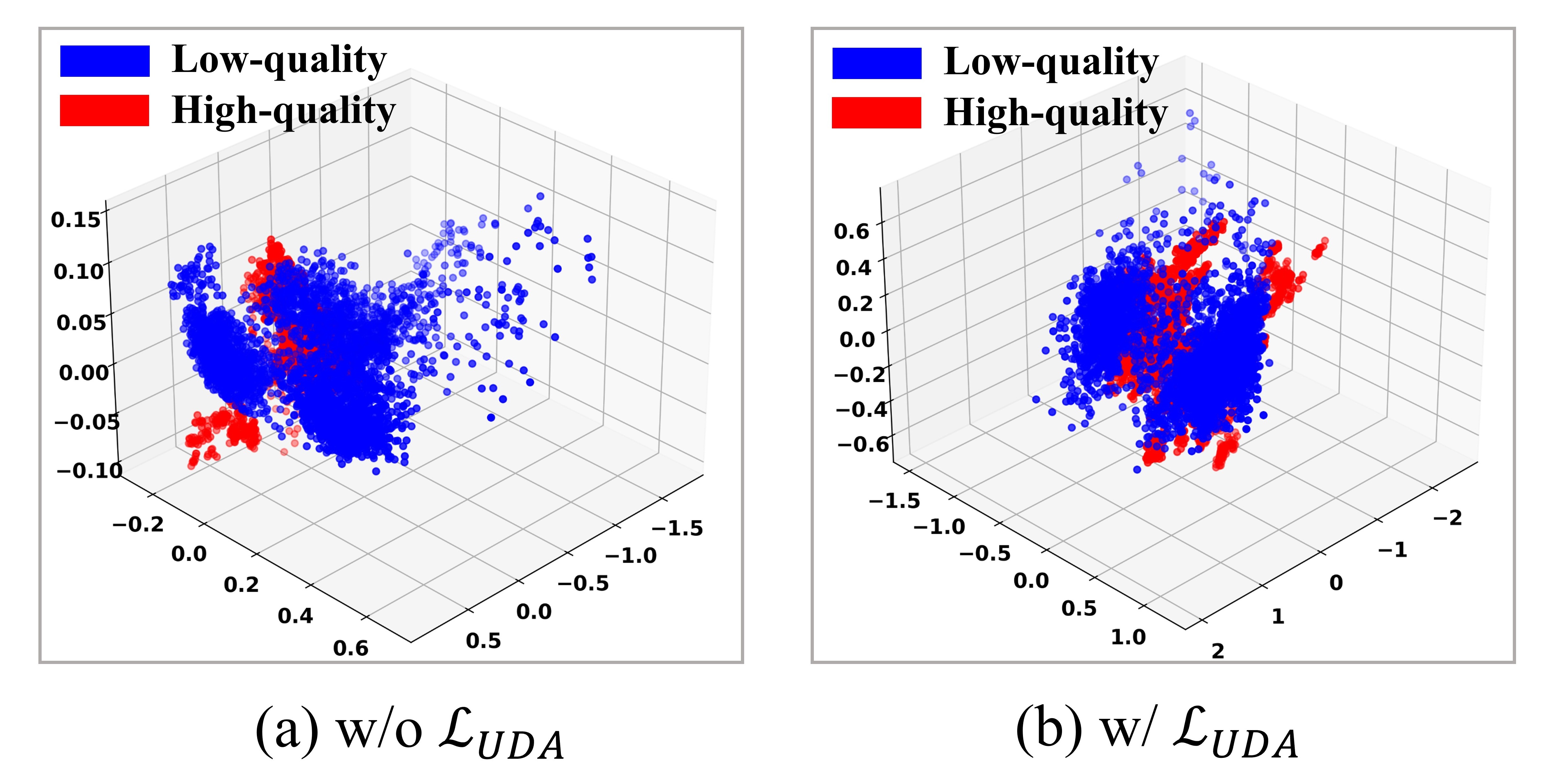}
	\caption{Monocular gaze feature distribution visualization. Different colors denote features from different samples.}
	\label{fig4}
\end{figure}
\begin{figure}
	\centering
	\includegraphics[width=\linewidth]{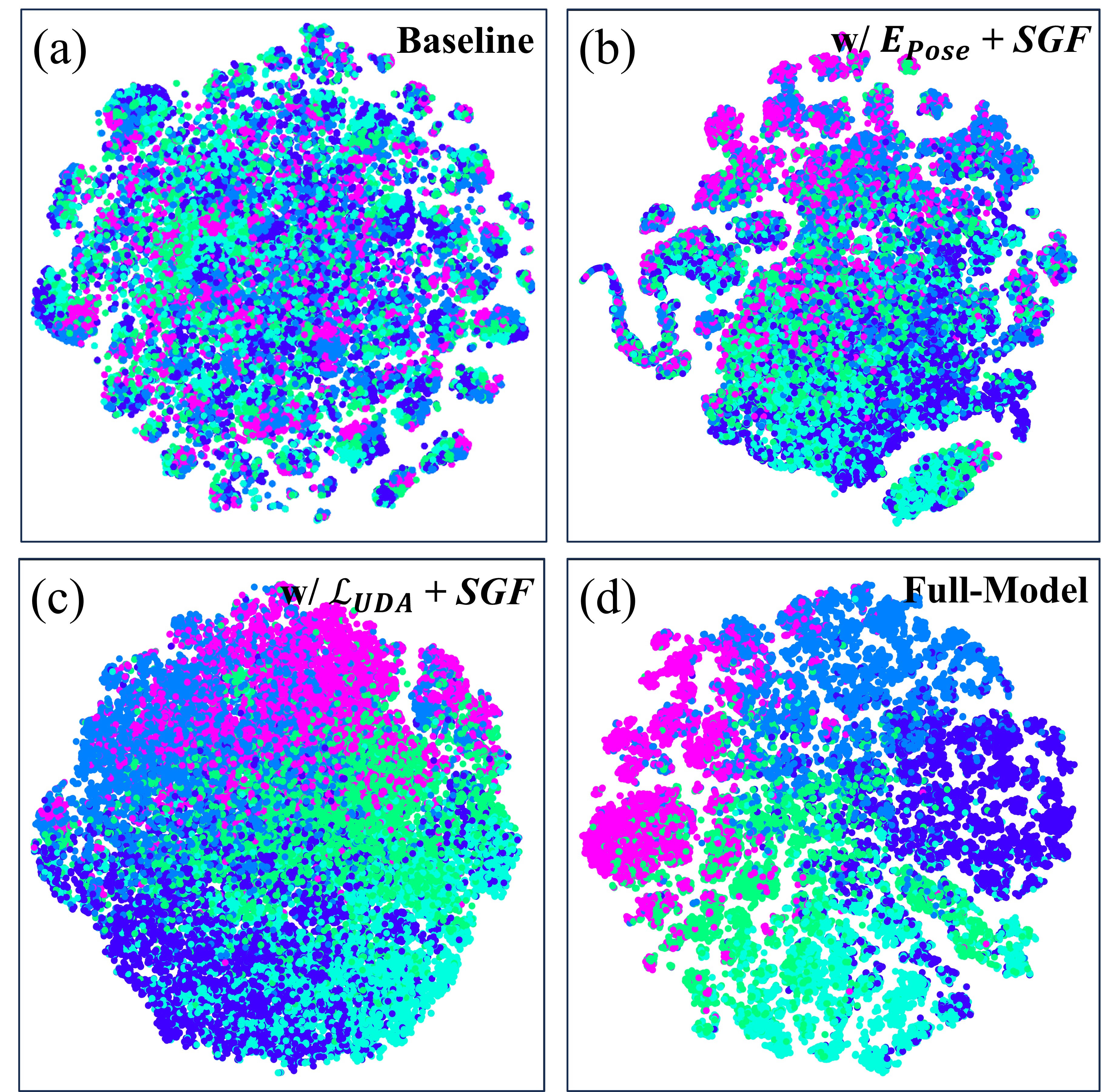}
	\caption{Visualization of the full-face gaze feature distribution. Different colors denotes different gaze directions and close gaze direction share similar colors.}
	\label{fig5}
\end{figure}

\subsubsection{Visualization Analysis} We visualize monocular gaze representation from UDAL to evaluate the effectiveness of the $\mathcal{L}_{\textit{UDA}}$, where we only remove it during the training, but low-quality and high-quality monocular images share an encoder. The features visualized results are shown in Fig~\ref{fig4}. It is obvious that removing $\mathcal{L}_{\textit{UDA}}$ leads to significant distribution gap in embedding space. Instead, with $\mathcal{L}_{\textit{UDA}}$, the features from high- and low-quality samples are aligned with consistent angles and scales, which implies the effectiveness of our domain-adaptation constraints.

To analyze the final gaze features of different variants, we further visualize the distribution of gaze features on domain generalization task $G \rightarrow M$ with $t$-SNE. Results are shown in Fig~\ref{fig5}, where the feature points with close gaze direction shared with similar colors. For the baseline model from Tab~\ref{tab3}, as shown in Fig~\ref{fig5}(a), the features with different gaze directions are mixed together and the feature cluster is quite dispersed. After injecting the $E_{\textit{Pose}}$ and SGF module into the baseline (shown in Fig~\ref{fig5}(b)), the feature distributions tend to become ordered, but they are still mixed. Then, we introduce $\mathcal{L}_{\textit{UDA}}$ and SGF module, the feature distribution becomes ordered with obvious cluster (shown in Fig~\ref{fig5}(c)). The best feature cluster appears in our full model, as shown in Fig~\ref{fig5}(d), the gaze direction and feature similarities have a strong correlation, which supports our insight, disentangling gaze representations from unconstrained facial images.

\section{Conclusion}
In this paper, we present a simple yet efficient Hybrid-domain Adaptative Representation Learning (HARL) framework, which is the first method to exploit hybrid-domain data to disentangle monocular gaze representation in an unsupervised domain adaptation manner, and further explores an efficient sparse-graph network to fuse head-pose representation, leading to an adaptative gaze representation learning architecture. The proposed HARL requires few parameters and computational costs and achieves significant gains in both in-domain and cross-domain evaluations. While many concrete implementations of the general idea, including utilizing powerful graph networks and fusion modules, are possible, we show that a simple design already achieves superior results, which provides new insights to solve robust gaze estimation in the wild.
\bigskip
\bibliography{aaai2026}

\end{document}